\documentclass{article}
\usepackage{amsmath}
\usepackage{amssymb}
\usepackage{amsthm}
\usepackage{natbib}
\usepackage{url}
\usepackage{a4wide}
\usepackage{parskip}
\usepackage{color}
\usepackage{hyperref}
\usepackage{tikz}
\usepackage{pgfplots}
\pgfplotsset{compat=newest}
\usepackage{booktabs}       
\usepackage{amsfonts}       
\usepackage{nicefrac}       
\usepackage{microtype}      
\usepackage{subcaption}

\usepackage[final]{nips_2017}

\renewcommand{\vec}[1]{\mathbf{#1}}
\newcommand{\trans}{^t}
\newcommand{\norm}[1]{\|#1\|}

\newcommand{\bn}{{Batch Normalization}}
\newcommand{\wn}{{Weight Normalization}}
\renewcommand{\ln}{{Layer Normalization}}

\newcommand{\tblref}[1]{Table~\ref{#1}}
\newcommand{\figref}[1]{Figure~\ref{#1}}

\newcommand{\secref}[1]{section~\ref{#1}}

\newskip\tblskipamount
\newcommand{\hlinetop}{\toprule}
\newcommand{\hlinemid}{\midrule}
\newcommand{\hlinebot}{\bottomrule}

\newcommand{\nonlin}{g}
\newcommand{\dnonlin}{\nonlin'(z)} 
\newcommand{\ddnonlin}{\nonlin''(z)} 
\newcommand{\z}{z}
\newcommand{\w}{\vec w}
\renewcommand{\v}{\vec v}

\newcommand{\g}{\vec g}

\newcommand{\E}{\text{E}}

\newcommand{\mom}{\rho}
\newcommand{\eps}{\epsilon}
\newcommand{\etaeff}{\eta_\text{eff}}
\newcommand{\etaopt}[1]{\eta^*}

\begin{document}

\title{$L_2$ Regularization versus Batch and Weight Normalization}
\author{Twan van Laarhoven\\
  Institute for Computer Science\\
  Radboud University\\
  Postbus 9010, 6500GL Nijmegen, The Netherlands \\
  \texttt{mail@twanvl.nl} \\
}
\maketitle





%


\begin{abstract}
  Batch Normalization is a commonly used trick to improve the training of deep neural networks.
  These neural networks use $L_2$ regularization, also called weight decay, ostensibly to prevent overfitting.
  However, we show that $L_2$ regularization has no regularizing effect when combined with normalization.
  Instead, regularization has an influence on the scale of weights, and thereby on the effective learning rate.
  We investigate this dependence, both in theory, and experimentally.
  We show that popular optimization methods such as ADAM only partially eliminate the influence of normalization on the learning rate.
  This leads to a discussion on other ways to mitigate this issue.
\end{abstract}

\section{Introduction}

Since its introduction in \citet{Ioffe2015BatchNorm}, Batch Normalization (BN) has quickly become a standard technique when training deep neural networks.
Batch Normalization attempts to solve the problem of covariate shift, that is, the change in the distribution of inputs to a unit as training progresses.
By normalizing these inputs to have zero mean and unit variance, training can be drastically sped up.

As the name suggests, Batch Normalization achieves this normalization by using the mean and variance of batches of training data.
More concretely, consider a single unit in a neural network. Its output is given by
\begin{align*}
  y_\text{NN}(X;\w,b) = \nonlin( X\w  + b ),
\end{align*}
where $\nonlin$ is a nonlinearity such as the rectified linear function (ReLU) or a sigmoid, $X$ is the input, and $\w$ and $b$ are the learned weights and bias.
In a a convolutional neural network, the weights can be shared with other units.

With Batch Normalization, instead the input to the nonlinearity is normalized,
\begin{align*}
  y_\text{BN}(X;\w,\gamma,\beta) = \nonlin\Bigl( \frac{X\w - \mu(X\w)}{\sigma(X\w)} \gamma + \beta \Bigr),
\end{align*}
where the mean $\mu$ and standard deviation $\sigma$ are computed given a batch $X$ of training data.
At test time, the values of $\mu$ and $\sigma$ are fixed.
The extra parameters $\gamma$ and $\beta$ are needed to still be able to represent all possible ranges of inputs to $\nonlin$.

More recently, \citet{Salimans2016weightNorm} introduced Weight Normalization (WN) as an alternative to batch normalization.
Here the normalization uses only the weights,
\begin{align*}
  y_\text{WN}(X;\w,\gamma,\beta) = \nonlin\Bigl( \frac{X\w}{\norm{\w}_2} \gamma + \beta \Bigr).
\end{align*}

Yet another normalization variant is Layer Normalization (LN), introduced by \citet{Ba2016LayerNorm}.
As in Batch Normalization, the mean and standard deviation are used to normalize the input to the nonlinearity, only instead of taking the statistics of a single unit over a whole batch of inputs, they are taken for a single input over all units in a layer,
\begin{align*}
  y_\text{LN}(\vec{x};W,\vec\gamma,\vec\beta) = \nonlin\Bigl( \frac{\vec{x} W - \mu(\vec{x} W)}{\sigma(\vec{x} W)} \vec{\gamma} + \vec{\beta} \Bigr).
\end{align*}

Networks using any of these normalization strategies are usually trained with variants of stochastic gradient descent (SGD), and using $L_2$ regularization of the weights.
The combination of $L_2$ regularization and gradient descend results in \emph{weight decay}, where in each update the weights are scaled by a factor slightly smaller than one.



However, each normalization strategies give functions that are invariant to scaling of the weights\footnote{LN is only invariant if the whole weight matrix for the layer is scaled uniformly.}, i.e. where
\begin{align*}
  y(X;\alpha \w,\gamma,\beta) = y(X;\w,\gamma,\beta).
\end{align*}
It is therefore surprising that $L_2$ regularization is still used.

In this paper we investigate the effects of $L_2$ regularization in combination with Batch, Weight and Layer Normalization.
We show that, as expected, there is no regularizing effect.
Rather, the `regularization' term strongly influences the learning rate.

\section{$L_2$ Regularization?}

The objective function of a neural network with $L_2$ is a combination of an unregularized objective and a regularization term,
\begin{align*}
  L_\lambda(\w) = L(\w) + \lambda \norm{\w}_2^2.
\end{align*}
The unregularized objective function depends on the weights only through the output of the unit,
\begin{align*}
  L(\w) = \sum_{i=1}^N \ell_i(y(X_i;\w,\gamma,\beta)),
\end{align*}
where $\ell_i$ is the loss with respect to the unit's output for a sample $i$.

But as we have seen, when using normalization, $y(X_i;\alpha\w,\gamma,\beta) = y(X_i;\w,\gamma,\beta)$, so
\begin{align*}
  L_\lambda(\alpha \w) = \sum_{i=1}^N \ell_i(y(X_i;\w,\gamma,\beta)) + \lambda \norm{\w}_2^2 = L_{\lambda\alpha^2}(\w).
\end{align*}
This means that the $L_2$ penalty term forces the weights to become small, as expected,
but that this has no regularizing effect of making the computed function simpler. The function $y$ is exactly the same, regardless of the scale of the weights.
And the parameter $\lambda$ has no impact on the optimum, since the weights can be scaled to compensate.



\section{Effect of the Scale of Weights on Learning Rate}

While the scale of weights $\w$ has no effect on the objective value, with first order optimization methods such as stochastic gradient, the scale of $\w$ does influence the updates that are performed.

With batch and layer normalization, the gradients are usually not propagated through the batch mean and standard deviation functions.
Hence the gradient of $y$ with respect to the weights is
\begin{align*}
  \nabla y_\text{BN}(X;\w,\gamma,\beta) =
  \frac{X}{\sigma(X\w)} \gamma \dnonlin,
\end{align*}
where $\z = (X\w - \mu(X\w))/\sigma(X\w) \gamma + \beta$ is the input to the nonlinearity.

When the weights are scaled by a factor of $\alpha$, the standard deviation and mean scale along. Hence we get that
\begin{align*}
  \nabla y_\text{BN}(X;\alpha\w,\gamma,\beta) =
  \nabla y_\text{BN}(X;\w,\gamma,\beta) / \alpha.
\end{align*}
So, while $y$ does not change when scaling the weights, its gradient does change!
The gradient of the objective scales similarly,
\begin{align*}
  \nabla L_\lambda(\alpha \w) &= \nabla L_{\lambda \alpha^2}(\w) / \alpha.
\end{align*}
%

For {\wn} the story is similar. In that case the gradient is
\begin{align*}
  \nabla y_\text{WN}(X;\alpha\w,\gamma,\beta) =
    \Bigl( \frac{X}{\norm{\w}} - \frac{\w X \w}{\norm{\w}^3} \Bigl) \gamma \dnonlin.
\end{align*}
And this again scales like $1/\alpha$ when the weights are scaled by $\alpha$.

\subsection{Effective Learning Rates}

\renewcommand{\time}[1]{_{#1}}
\renewcommand{\next}{\time{t+1}}
\newcommand{\cur}{\time{t}}
\newcommand{\alltime}{\time{t}}

A typical update with Stochastic Gradient Descend (SGD) looks like
\begin{align*}
  \w\time{t+1} \gets \w\time{t} - \eta \nabla L_\lambda(\w\time{t})
\end{align*}
Now imagine that $\w\time{t} = \alpha \w'\time{t}$.
Then we can write the SGD update as
\begin{align*}
  \alpha\w'\time{t} \gets \alpha\w'\time{t} - \eta/\alpha \nabla L_{\lambda\alpha^2}(\w\time{t}').
\end{align*}
But this is just an SGD update of $\w'$ with a different learning rate and different amount of regularization,
\begin{align*}
  \w' \gets \w' - \eta/\alpha^2 \nabla L_{\lambda\alpha^2}(\w').
\end{align*}

We can imagine all updates being performed on a vector of weights $\w'$ with norm 1, so $\alpha=\norm{w}_2$ is the scale of the weights.
The \emph{effective learning rate} in terms of the normalized $\w'$ is then $\etaeff = \eta/\norm{w}_2^2$.
Which means that by decreasing the scale of the weights, weight decay \emph{increases} the effective learning rate.
This is contrary to the intuition that regularization should result in a more stable model.


In practice, people use cross-validation to pick parameters such as the learning rate $\eta$ and the amount of regularization $\lambda$.
Suppose that for a given value of $\lambda$ the optimal learning rate is $\etaopt{\lambda}$.
Then the by scaling the weights by $\sqrt{\lambda'/\lambda}$ we see that the same effective updates are performed for $\lambda'$ and $\lambda/\lambda' \etaopt{\lambda}$.
Hence $\etaopt{\lambda'} = \lambda/\lambda' \etaopt{\lambda}$ is the optimal learning rate for $\lambda'$.
This implies that $\etaopt{\lambda}$ must be proportional to $1/\lambda$.

We can now also answer the question of why $L_2$ regularization is still beneficial when training neural networks with {\bn}:
If no regularization is used the weights can grow unbounded, and the effective learning rate goes to $0$.
\section{Effect of Regularization on the Scale of Weights}
\label{sec:effect-on-scale}

In the previous section we discussed how scaling the weights affects the effective learning rate.
This still requires that the weights are scaled. So in this section we will look at how the scale of the weights is affected by $L_2$ regularization.

During training the scale of the weights will change. The gradients of the loss function, and in particular the stochastic nature of these gradients, will cause the norm of the weights to grow, while the regularization term causes the weights to shrink. As a result, the effective learning rate also changes over time as training progresses.

\newcommand{\stddev}{\sigma}
After convergence the expected gradient is $0$,
so as a simple model of this process, we can pretend that the input $X$ and the gradients with respect to the output $\nabla L(y)$ are drawn from some simple probability distribution. Say, $X \sim \mathcal{N}(0,\stddev_X^2)$ and $\nabla L(y) \sim \mathcal{N}(0,\stddev_\nabla^2)$.

Then when using {\bn}, we get that $\E[\nabla L(\w\cur)]=0$ while $\E[\nabla L(\w\cur)^2] = \stddev_\nabla^2 \gamma^2 \dnonlin / \|\w\cur\|_2^2$.
Furthermore, if the nonlinearity is a rectifier, then its gradient $\dnonlin$ is $0$ or $1$ with equal probability, and we get that $\E[\nabla L(\w\cur)^2] = \stddev_\nabla^2 \gamma^2 / (2 \|\w\cur\|^2) $.
Hence the expected norm of $\w\next$ is
\begin{equation*}
  \E[\|\w\next\|_2^2] = (1-\eta\lambda)^2\|\w\cur\|_2^2 + \frac{\eta^2 \gamma^2 \stddev_\nabla^2}{2\|\w\cur\|_2^2}.
\end{equation*}

Now suppose that the training has converged, i.e. that $\|\w\next\| = \|\w\cur\|$. We then find that the norm of the weights is
\begin{equation*}
  \|\w\cur\| = \sqrt[4]{\frac{\eta^2\gamma^2\stddev_\nabla^2}{ 4\eta\lambda - 2\eta^2\lambda^2 }}.
\end{equation*}
If $\eta\lambda$ is sufficiently small then this is approximately
\begin{equation*}\|\w\cur\| \approx O(\sqrt[4]{\eta/\lambda}).\end{equation*}


Combining this with the effective learning rate from the previous section, we get that the effective learning rate in terms of $\eta$ and $\lambda$ is approximately $\eta_\text{eff} = \sqrt{\eta/\lambda}$.
Picking $\eta \propto 1/\lambda$ as previously discussed results in a constant effective learning rate.

\section{Other Update Rules}

In practice, plain stochastic gradient updates are almost never used to train neural networks.
Below we discuss several other update rules, and the effective learning rates when weights are scaled.
These results are summarized in \tblref{tbl:update-rules}.

\begin{table}[t]
  \caption{Effective learning rate and scaling of weights for various update rules.}
  \label{tbl:update-rules}
  \centering
  \small
  \begin{tabular}{lccc}
    \hlinetop
    {Update rule} &
    {Effective} &
    {Weight scale} & 
    {Effective}
    \\
    &
    {learning rate}&
    &
    {learning rate}\\
    &
    $\etaeff$&
    $\norm{\w}_2$&
    $\etaeff$\\
    \hlinemid
    SGD
       & $\eta/\norm{\w}_2^2$  & $O(\sqrt[4]{\eta/\lambda})$ & $O(\sqrt{\eta/\lambda})$ \\
    Momentum
       & $\eta/\norm{\w}_2^2$  & $O(\sqrt[4]{\eta/\lambda})$ & $O(\sqrt{\eta/\lambda})$ \\
    RMSProp
       & $\eta/\norm{\w}_2$  & $O(\sqrt[3]{\eta/\lambda})$ & $O(\sqrt[3]{\eta^2/\lambda})$ \\
    ADAM
       & $\eta/\norm{\w}_2$  & $O(\sqrt[3]{\eta/\lambda})$ & $O(\sqrt[3]{\eta^2/\lambda})$ \\
    Normalized SGD
       & $\eta$  & $1$ & $\eta$ \\
    Newton
       & $\eta$ & $\to 0$ & $\eta$ \\
    \hlinebot
  \end{tabular}
\end{table}

\subsection{Momentum}

A common extension to SGD is to apply momentum \citep{Sutskever2013importance}, to obtain updates such as
\begin{align*}
  \label{eq:momentum}
  \v\next &\gets \mom \v\cur - \eta \nabla L_\lambda(\w\cur) \\
  \w\next &\gets \w\cur - \v\next
\end{align*}
or
\begin{align*}
  \v\next &\gets \mom \v\cur - \eta \nabla L_\lambda(\w\cur + \mom\v\cur) \\
  \w\next &\gets \w\cur - \v\next
\end{align*}
Repeating the above analysis, we see that, as with SGD, scaling the weights by $\alpha$ corresponds to performing updates with learning rate $\eta/\alpha^2$, hence the effective learning rate is $\eta/\norm{w}_2^2$.

Similarly, it can be shown that the norm of the weights converges to the order of $\sqrt[4]{\nicefrac\eta{(1-\rho)\lambda}}$.

\subsection{RMSProp}

The problem that stochastic gradient methods depend on scaling of the objective is well known.
Indeed, we observe a similar effect if the objective is scaled by a constant to obtain $\alpha L(\w)$.
To address this scaling problem, methods have been developed that automatically adapt the learning rate based on the norm of the gradients.

One of these methods, RMSProp \citep{TielemanHinton2012rmsprop}, uses the update rule,
\begin{align*}
  \g\next &\gets \rho \g\cur + (1-\rho) \nabla L_\lambda(\w\cur)^2 \\
  \w\next &\gets \w\cur - \eta  \frac{\nabla L_\lambda(\w\cur)}{\sqrt{\g\next + \eps}}.
\end{align*}
Now suppose that $\w\time{t}=\alpha \w'\time{t}$ and $\g\time{t} = \g'\time{t} / \alpha^2$. Then this update is
\begin{align*}
  \frac{\g'\next}{\alpha^2} &\gets \rho \frac{\g'\cur}{\alpha^2} + (1-\rho) \frac{L_{\lambda\alpha^2}(\w'\cur)^2}{\alpha^2} \\
  \alpha \w'\next &\gets \alpha \w'\cur - \eta \frac{\nabla L_{\lambda\alpha^2}(\w'\cur)}{\alpha \sqrt{\g'\next/\alpha^2 + \eps}},
\end{align*}
which is equivalent to RMSProp updates with learning rate $\eta/\alpha$, and $\eps\alpha^2$ instead of $\eps$.
\begin{align*}
  \g'\next &\gets \rho \g'\cur + (1-\rho) \nabla L_\lambda(\w'\cur)^2 \\
  \w'\next &\gets \w'\cur - \eta/\alpha  \frac{\nabla L_\lambda(\w'\cur)}{\sqrt{\g'\next + \eps\alpha^2}}.
\end{align*}
The effective learning rate is therefore $\eta / \norm{\w}_2$.
In terms of the regularization parameter, we get that for RMSProp the optimal learning rate is $\etaopt{\lambda} \propto 1/\sqrt{\lambda}$.

So, while this situation is different, and arguably improved compare to SGD; using RMSprop does not eliminate the dependence of the learning rate on the scale of the weights is not completely resolved.
%

If we look at the relation between the learning rate and the norm of the weights, we use the same model as in \secref{sec:effect-on-scale}.
So, suppose that training has converged, that is, $\norm{w\next}= \norm{w\cur}$.
Then $\E[\g\next = \stddev_\nabla^2 \gamma^2 / (2 \norm{\w\cur}_2^2]$.
And if $\eps$ is sufficiently small, we get that 
\begin{align*}
  \E[\norm{\w\next}_2^2] = \Bigl( 1 - \frac{\eta\lambda \norm{\w\cur}_2}{\stddev_\nabla \gamma} \Bigr)^2 \norm{\w\cur}_2^2 + \frac{\eta^2}{2}.
\end{align*}
Equating $\norm{\w\next}= \norm{\w\cur}$ we find that in expectation
\begin{align*}
  \norm{\w}_2 = O\Bigl(\sqrt[3]{\frac{\eta \stddev_\nabla \gamma}{\lambda}}\Bigr).
\end{align*}
So the weights scale with the cube root of $\eta/\lambda$.
We can thus write the effective learning rate 
as  $\etaeff = \sqrt[3]{\eta^2 / \lambda}$.

\subsection{ADAM}
ADAM \citep{Kingma2014adam} is a scheme that combines ideas from RMSProp with momentum.
Again using the norm of the gradient to scale the learning rate.
The updates rule is
\begin{align*}
  \v\next &\gets \rho_1 \v\cur + (1-\rho_1) \nabla L_\lambda(\w\cur) \\
  \g\next &\gets \rho_2 \g\cur + (1-\rho_2) \nabla L_\lambda(\w\cur)^2 \\
  \w\next &\gets \w\cur - \eta \v\next /(1-\rho_1) /(\sqrt{\g\next/(1-\rho_2)} + \eps).
\end{align*}
This is equivalent to updates of $\w\time{t}=\alpha \w'\time{t}$, $\v\time{t}=\v'\time{t}/\alpha'$, $\g\time{t}=\g'\time{t}/\alpha^2$ with $\eta'=\eta/\alpha$ and $\eps'=\alpha\eps$.
Again, the effective learning rate is $\eta / \norm{\w}_2$.

\subsection{Second Order Methods}
The problems with gradient descend's dependence on scaling of the parameters and objective function stems from the fact that it is a first order method. And so, as the scale of the parameters decreases, the gradient increases instead of decreasing.
Both RMSProp and ADAM try to use the to correct for this by using the norm of the gradient, but as we have shown, they are only partially successful.
A true solution would be to use second derivatives.

In non-stochastic optimization the standard second order optimization method is Newton's method, whose updates look like
\begin{align*}
  \w\next &\gets \w\cur - \eta \frac{\nabla L_\lambda(\w\cur) }{ \nabla^2 L_\lambda(\w\cur)},
\end{align*}
where $\nabla^2$ denotes the second derivative or Hessian matrix.

To investigate how Newton's method is affected by scaling of $\w$,
we first need to compute the second derivative of $L$ when using {\bn},
\begin{align*}
  \nabla^2 y_\text{BN}(X;\w,\gamma,\beta) =
  \frac{X\trans X}{\sigma(X\w)^2} \gamma^2 \ddnonlin.
\end{align*}
Which scales like $1/\alpha^2$ when the weights are scaled by $\alpha$.
That means that
\begin{align*}
  \nabla^2 L_\lambda(\w\cur) = \nabla^2 L_{\lambda\alpha^2}(\w\cur) / \alpha^2.
\end{align*}
The same holds for {\wn} and {\ln}.

Going back to Newton's method, and supposing that $\w\time{t}=\alpha\w'\time{t}$, we see that
\begin{align*}
  \alpha\w'\next &\gets \alpha\w'\cur - \eta \frac{\nabla L_{\lambda\alpha^2}(\w'\cur)/\alpha }{ \nabla^2 L_{\lambda\alpha^2}(\w'\cur)/\alpha^2},
\end{align*}
which is equivalent to Newton updates of $\w'$,
\begin{align*}
  \w'\next &\gets \w'\cur - \eta \frac{\nabla L_{\lambda\alpha^2}(\w'\cur)}{ \nabla^2 L_{\lambda\alpha^2}(\w'\cur)}.
\end{align*}
So, for Newton's method, scaling the weights does not affect the learning rate.
In other words, the effective learning rate is $\etaeff = \eta$.

However, Newton's method is not practical for training neural networks, because the Hessian matrix is too large to compute explicitly.
An alternative is Hessian Free optimization \citep{martens2010hessianFree}, where only multiplication with the Hessian is needed.
Like Newton's method, Hessian Free optimization is not affected by the scaling of weights.



\subsection{Normalizing Weights}
A brute force approach to avoid the interaction between the regularization parameter and the learning rate is to fix the scale of the weights.
We can do this by rescaling the $\w$ to have norm $1$:
\begin{align*}
  \tilde{\w}\next &\gets \w\cur - \eta \nabla L_\lambda(\w\cur)\\
  \w\next &\gets \tilde{\w}\next / \norm{\tilde{\w}\next}_2.
\end{align*}
With this change, the scale of the weights obviously no longer changes during training, and so the effective rate no longer depends on the regularization parameter $\lambda$.
Note that this weight normalizing update is different from {\wn}, since there the norm is taken into account in the computation of the gradient, but is not otherwise fixed.

%
%

\section{Experimental Validation}

\begin{figure}
  \centering
  \begin{subfigure}{.41\linewidth}
    \centering
    \vspace*{2.2mm}%
    \begin{tikzpicture}
      \begin{loglogaxis}[
          xlabel={regularization ($\lambda$)},
          ylabel={learning rate ($\eta$)},
          view={0}{90},
          colormap={myg}{[0.05cm] rgb(0cm)=(0,0,0); rgb(0.35cm)=(0.1,0.5,0.2); rgb(0.65cm)=(0.7,1,0.8); rgb(1cm)=(1,1,1);  },
          width=58mm,
          height=60mm,
          point meta min=-1,point meta max=2,
          xmin=0.0000032,xmax=3.1,
          ymin=0.000070,ymax=7,
          shader=flat corner,
          ]
        \addplot3[surf,mesh/ordering=y varies,draw=mapped color!80!gray] file {weight_norm_nesterov_weights_only.tex};
      \end{loglogaxis}
    \end{tikzpicture}
    \caption{Nesterov momentum}
  \end{subfigure}%
  \begin{subfigure}{.59\linewidth}
    \centering
    \begin{tikzpicture}
      \begin{loglogaxis}[
          xlabel={regularization ($\lambda$)},
          ylabel={learning rate ($\eta$)},
          view={0}{90},
          colormap={myg}{[0.05cm] rgb(0cm)=(0,0,0); rgb(0.35cm)=(0.1,0.5,0.2); rgb(0.65cm)=(0.7,1,0.8); rgb(1cm)=(1,1,1);  },
          width=58mm,
          height=60mm,
          point meta min=-1,point meta max=2,
          xmin=0.0000032,xmax=3.1, ymin=0.000070,ymax=7,
          shader=flat corner,
          colorbar,
          colorbar style={yticklabel=$10^{\pgfmathparse{\tick}\pgfmathprintnumber\pgfmathresult}$}
          ]
        \addplot3[surf,mesh/ordering=y varies,draw=mapped color!80!gray] file {weight_norm_adam_weights_only.tex};
      \end{loglogaxis}
    \end{tikzpicture}%
    \caption{ADAM}
  \end{subfigure}
  \caption{
    Norm of the first layer weights a function of learning rate and regularization.
  }
  \label{fig:norm-weights}
\end{figure}
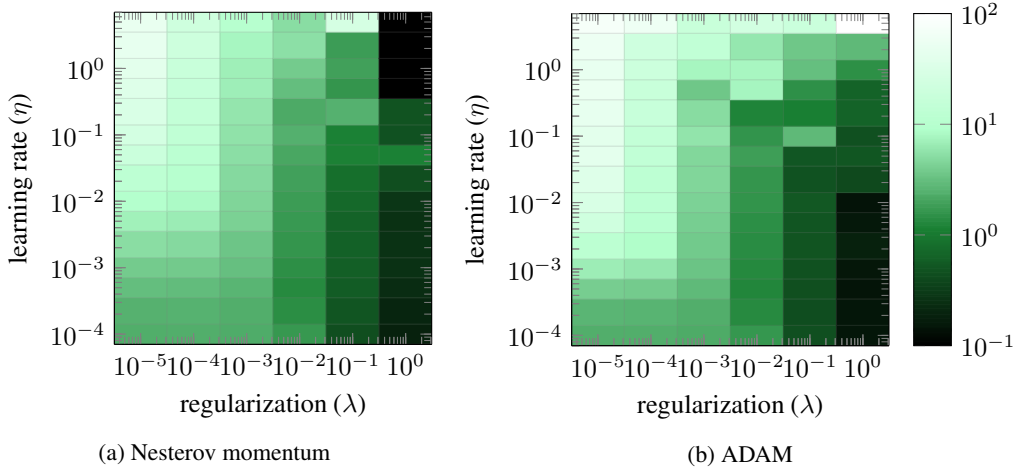

To demonstrate the relation between regularization and learning rates, we have performed a series of experiments using the \textsc{cifar10} dataset \citep{Krizhevsky2009cifar}.
We augment the training data by translating and horizontally flipping images.
We use a simple network with 4 convolutional layers, and use {\bn}.


We train for 100 epochs, and decrease the learning rate by a factor of 10 after 50 epochs and again after 80 epochs.
For SGD with Nesterov momentum we use a decay rate of $\mom=0.9$, for RMSProp and ADAM we use $\rho=0.9$ and for ADAM $\rho_1=0.9$ and $\rho_2=0.999$. We use a batch size of 128 in all experiments.

In \figref{fig:norm-weights} we plot at the norm of the weights as a function of the learning rate $\eta$ and the regularization $\lambda$.
The effect of $\lambda$ s seen most clearly for high $\eta$, while the effect of $\eta$ is seen best with low $\lambda$, that is, when the effective learning rate is high.
In these regions training has time to converge, and the norm of the weights resembles the theoretical $O(\sqrt[4]{\eta/\lambda})$ for Nesterov momentum momentum and  $O(\sqrt[3]{\eta/\lambda})$ for ADAM.

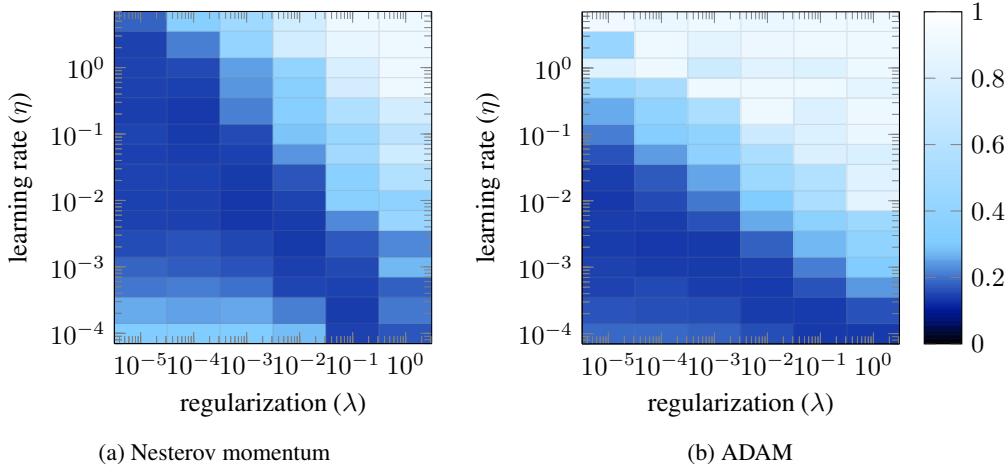
\begin{figure}
  \centering%
  \begin{subfigure}{.41\linewidth}
    \centering
    \vspace*{2.2mm}%
    \begin{tikzpicture}
      \begin{loglogaxis}[
          xlabel={regularization ($\lambda$)},
          ylabel={learning rate ($\eta$)},
          view={0}{90},
          colormap={my}{[0.05cm] rgb(0cm)=(0,0,0); rgb(0.10cm)=(0.0,0.1,0.6); rgb(0.30cm)=(0.5,0.8,1); rgb(1cm)=(1,1,1);  },
          width=58mm,
          height=60mm,
          point meta min=0,point meta max=1,
          xmin=0.0000032,xmax=3.1, ymin=0.000070,ymax=7,
          shader=flat corner,
          ]
        \addplot3[surf,mesh/ordering=y varies,draw=mapped color!80!gray] file {test_error_nesterov_weights_only.tex};
      \end{loglogaxis}
    \end{tikzpicture}%
    \caption{Nesterov momentum}
  \end{subfigure}%
  \begin{subfigure}{.59\linewidth}
    \centering
    \begin{tikzpicture}
      \begin{loglogaxis}[
          xlabel={regularization ($\lambda$)},
          ylabel={learning rate ($\eta$)},
          view={0}{90},
          colormap={my}{[0.05cm] rgb(0cm)=(0,0,0); rgb(0.10cm)=(0.0,0.1,0.6); rgb(0.30cm)=(0.5,0.8,1); rgb(1cm)=(1,1,1);  },
          width=58mm,
          height=60mm,
          point meta min=0,point meta max=1,
          xmin=0.0000032,xmax=3.1, ymin=0.000070,ymax=7,
          shader=flat corner,
          colorbar
          ]
        \addplot3[surf,mesh/ordering=y varies,draw=mapped color!80!gray] file {test_error_adam_weights_only.tex};
      \end{loglogaxis}
    \end{tikzpicture}%
    \caption{ADAM}
  \end{subfigure}
  \caption{
    Test set error as a function of learning rate and regularization.
  }
  \label{fig:test-error}
\end{figure}

Next, in \figref{fig:test-error} we look at the error on the test set as a function of the learning rate $\eta$ and the regularization $\lambda$. We know that for Nesterov momentum the effective learning rate is $\sqrt{\eta/\lambda}$, and for ADAM the effective learning rate is $\sqrt[3]{\eta^2/\lambda}$. Along the diagonals where the effective learning rate is constant, the test error is also roughly constant.

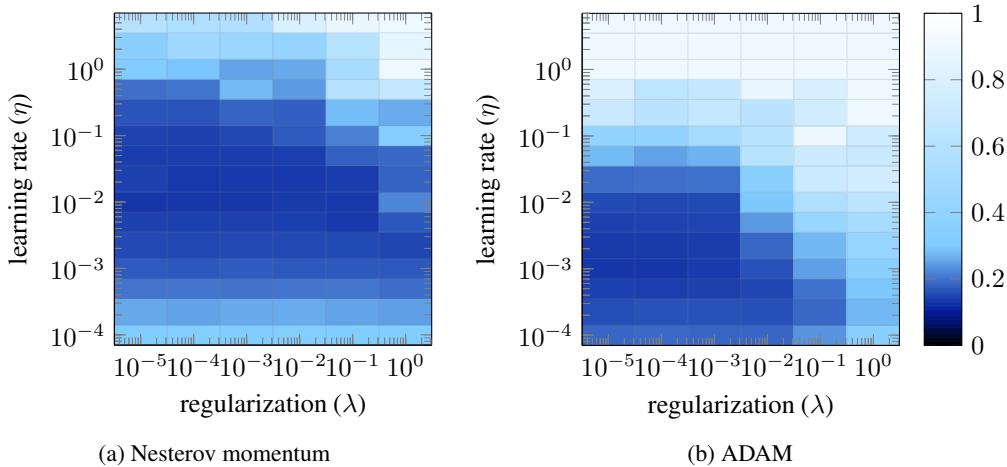
\begin{figure}
  \centering%
  \begin{subfigure}{.41\linewidth}
    \centering
    \vspace*{2.2mm}%
    \begin{tikzpicture}
      \begin{loglogaxis}[
          xlabel={regularization ($\lambda$)},
          ylabel={learning rate ($\eta$)},
          view={0}{90},
          colormap={my}{[0.05cm] rgb(0cm)=(0,0,0); rgb(0.10cm)=(0.0,0.1,0.6); rgb(0.30cm)=(0.5,0.8,1); rgb(1cm)=(1,1,1);  },
          width=58mm,
          height=60mm,
          point meta min=0,point meta max=1,
          xmin=0.0000032,xmax=3.1, ymin=0.000070,ymax=7,
          shader=flat corner,
          ]
        \addplot3[surf,mesh/ordering=y varies,draw=mapped color!80!gray] file {test_error_normalize_nesterov_weights_only.tex};
      \end{loglogaxis}
    \end{tikzpicture}%
    \caption{Nesterov momentum}
  \end{subfigure}%
  \begin{subfigure}{.59\linewidth}
    \centering
    \begin{tikzpicture}
      \begin{loglogaxis}[
          xlabel={regularization ($\lambda$)},
          ylabel={learning rate ($\eta$)},
          view={0}{90},
          colormap={my}{[0.05cm] rgb(0cm)=(0,0,0); rgb(0.10cm)=(0.0,0.1,0.6); rgb(0.30cm)=(0.5,0.8,1); rgb(1cm)=(1,1,1);  },
          width=58mm,
          height=60mm,
          point meta min=0,point meta max=1,
          xmin=0.0000032,xmax=3.1, ymin=0.000070,ymax=7,
          shader=flat corner,
          colorbar
          ]
        \addplot3[surf,mesh/ordering=y varies,draw=mapped color!80!gray] file {test_error_normalize_rmsprop_weights_only.tex};
      \end{loglogaxis}
    \end{tikzpicture}%
    \caption{ADAM}
  \end{subfigure}
  \caption{
    Test set error when normalizing the weights to have norm 1.
  }
  \label{fig:test-error-normalized}
\end{figure}

When normalizing the weights after each update we see that the effect of the regularization parameter almost disappears, see \figref{fig:test-error-normalized}. Only for large $\lambda$ is the test error worse, because then $\eta\lambda\w$ constitutes a significant part of the gradient.

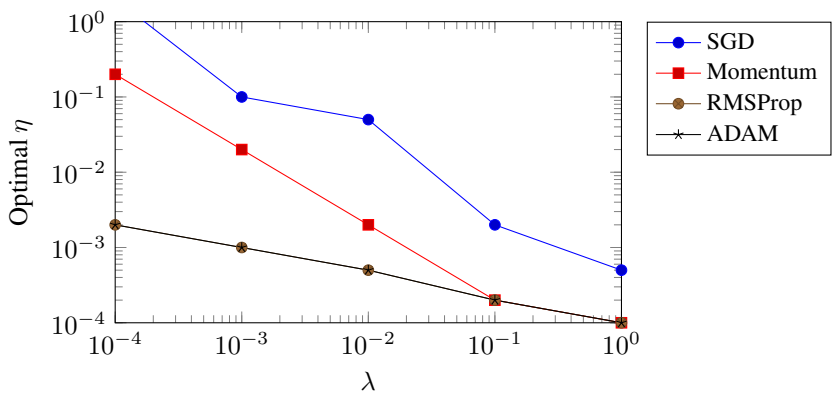
\begin{figure}
  \centering
  \begin{tikzpicture}
    \begin{loglogaxis}
      [xlabel=$\lambda$
      ,ylabel=Optimal $\eta$
      ,width=0.595\linewidth
      ,height=0.4\linewidth
      ,minor tick num=1
      ,xmin=0.0001,xmax=1,ymin=0.0001,ymax=1,enlargelimits=false
      ,xlabel near ticks
      ,legend style={cells={anchor=west},anchor=north west,at={(1.05,1.0)},font=\small}
      ]
      \addplot+[] file {best_learning_rate_sgd_weights_only.tex};
      \addlegendentry{SGD}
      \addplot+[] file {best_learning_rate_nesterov_weights_only.tex};
      \addlegendentry{Momentum}
      \addplot+[] file {best_learning_rate_rmsprop_weights_only.tex};
      \addlegendentry{RMSProp}
      \addplot+[] file {best_learning_rate_adam_weights_only.tex};
      \addlegendentry{ADAM}
    \end{loglogaxis}
  \end{tikzpicture}
  \caption{Optimal learning rate determined by cross-validation, as a function of the regularization parameter $\lambda$.}
  \label{fig:learning-rate}
\end{figure}

In \figref{fig:learning-rate}
we plot the optimal learning rate determined by cross-validation as a function of $\eta$.
The results closely match the theory: for SGD and Nesterov momentum updates we see an inverse linear relationship, $\etaopt{\lambda} \propto 1/\lambda$, while for RMSProp and ADAM the relation is $\etaopt{\lambda} \propto 1/\sqrt \lambda$.

\section{Discussion}

Normalization, either {\bn}, {\ln}, or {\wn} makes the learned function invariant to scaling of the weights $\w$.
This scaling is strongly affected by regularization.
We know of no first order gradient method that can fully eliminate this effect.
However, a direct solution of forcing $\norm{\w}=1$ solves the problem. By doing this we also remove one hyperparameter from the training procedure.

As noted by \citet{Salimans2016weightNorm}, the effect of weight and batch normalization on the effective learning rate might not necessarily be bad. 
If no regularization is used, then the norm of the weights tends to increase over time, and so the effective learning rate decreases. Often that is a desirable thing, and many training methods lower the learning rate explicitly. However, the decrease of effective learning rate can be hard to control, and can depend a lot on initial steps of training, which makes it harder to reproduce results.


With batch normalization we have added two additional parameters, $\gamma$ and $\beta$, and it of course makes sense to also regularize these.
In our experiments we did not use regularization for these parameters, though preliminary experiments show that regularization here does not affect the results.
This is not very surprising, since with rectified linear activation functions, scaling of $\gamma$ also has no effect on the function value in subsequent layers. So the only parameters that are actually regularized are the $\gamma$'s for the last layer of the network.

\section*{Acknowledgements}

This work has been partially funded by the Netherlands Organization for Scientific Research (NWO) within the EW TOP Compartiment 1 project 612.001.352.

\bibliographystyle{icml2017}
\bibliography{deep-learning}

\end{document}